\documentclass[sigconf]{acmart}
\def\BibTeX{{\rm B\kern-.05em{\sc i\kern-.025em b}\kern-.08emT\kern-.1667em\lower.7ex\hbox{E}\kern-.125emX}}

\usepackage{makecell}

\graphicspath{ {figures/} }

\def\acmBooktitle#1{\gdef\@acmBooktitle{#1}}
\acmBooktitle{Proceedings of \acmConference@name
       \ifx\acmConference@name\acmConference@shortname\else
         \ (\acmConference@shortname)\fi}

\title{Entertaining and Opinionated but Too Controlling: \\ A Large-Scale User Study  of  an Open Domain Alexa Prize System}
\renewcommand{\shorttitle}{User Experiences Of Automated Social Conversations}

\author{Kevin K. Bowden}
\orcid{1-5208-1170}
\affiliation{%
  \institution{NLDS$^{*}$\authornote{Natural Language and Dialogue Systems} Laboratory\\
  University of California Santa Cruz\\}
}
\email{kkbowden@ucsc.edu}

\author{Jiaqi Wu}
\affiliation{%
  \institution{NLDS$^{*}$ Laboratory\\
  University of California Santa Cruz\\}
}
\email{jwu64@ucsc.edu}

\author{Wen Cui}
\affiliation{%
  \institution{NLDS$^{*}$ Laboratory\\
  University of California Santa Cruz\\}
}
\email{wcui7@ucsc.edu}

\author{Juraj Juraska}
\affiliation{%
  \institution{NLDS$^{*}$ Laboratory\\
  University of California Santa Cruz\\}
}
\email{jjuraska@ucsc.edu}

\author{Vrindavan Harrison}
\affiliation{%
  \institution{NLDS$^{*}$ Laboratory\\
  University of California Santa Cruz\\}
}
\email{vharriso@ucsc.edu}

\author{Brian Schwarzmann}
\affiliation{%
  \institution{NLDS$^{*}$ Laboratory\\
  University of California Santa Cruz\\}
}
\email{brschwar@ucsc.edu}

\author{Nicholas Santer}
\affiliation{%
  \institution{Re-Embodied Cognition Laboratory\\
  University of California Santa Cruz\\}
}
\email{nsanter@ucsc.edu}

\author{Steve Whittaker}
\affiliation{%
  \institution{HCI$^{\dagger}$ \authornote{Human Computer Interaction} Laboratory\\
  University of California Santa Cruz\\}
}
\email{swhittak@ucsc.edu}

\author{Marilyn Walker}
\affiliation{%
  \institution{NLDS$^{*}$ Laboratory\\
  University of California Santa Cruz\\}
}
\email{mawalker@ucsc.edu}

\renewcommand{\shortauthors}{Bowden and Wu, et al.}

\date{}

\begin{document}

\makeatletter
\renewcommand{\@copyrightowner}{Copyright held by the owner/author(s). Publication rights licensed to Association for Computing Machinery.}
\makeatother
\copyrightyear{2019} 
\acmYear{2019} 
\acmConference[CUI 2019]{1st International Conference on Conversational User Interfaces}{August 22--23, 2019}{Dublin, Ireland}
\acmBooktitle{1st International Conference on Conversational User Interfaces (CUI 2019), August 22--23, 2019, Dublin, Ireland}
\acmPrice{15.00}
\acmDOI{10.1145/3342775.3342792}
\acmISBN{978-1-4503-7187-2/19/08}
\setcopyright{acmlicensed}

\begin{abstract}

Conversational systems typically focus  on functional tasks such as scheduling appointments or creating todo lists. Instead we design and evaluate SlugBot (SB), one of 8 semifinalists in the 2018 Alexa Prize, whose goal is  to support casual open-domain social interaction. This novel application requires both broad topic coverage and engaging interactive skills. We developed a new technical approach to meet this demanding situation by crowd-sourcing novel content and introducing  playful conversational strategies based on storytelling and games. We collected over 10,000 conversations during August 2018 as part of the Alexa Prize competition. We  also conducted an in-lab follow-up qualitative evaluation. Overall users found SB  moderately engaging; conversations averaged 3.6 minutes and involved 26 user turns. However, users reacted very differently to different conversation subtypes. Storytelling and games were evaluated positively; these were seen as entertaining with predictable interactive structure. They also led users to impute personality and intelligence to SB. In contrast, search and general Chit-Chat induced coverage problems; here users found it hard to infer what topics SB could understand, with these conversations seen as being too system-driven. Theoretical and design implications suggest a move away from conversational systems that simply provide factual information. Future systems should be designed to have their own opinions with personal stories to share, and SB provides an example of how we might achieve this. 

\end{abstract}



\maketitle

\section{Introduction}

Human conversation has long been an enticing metaphor for human
computer interaction, with arguments offered that interacting with computers should resemble a natural interaction between people. But although conversational agents have existed for many years \cite{Schmandt85,WS89,CassellThorisson99,Stent01,Stallard00}, successful systems have had to limit themselves  to particular tasks and constrained system functionality.
However, the recent  resurgence of interest in  conversational systems, occasioned by  a new generation of commercial personal assistants,   has led to an interest in {\bf open domain} conversation, which seems feasible for the first time due to vastly improved speech recognition, search, and natural language understanding. In the main however, current deployed systems still focus on the execution of practical tasks. 

This paper describes how we designed and deployed a novel open-domain social conversational system, SlugBot (SB), one of eight conversational systems competing in the 2018 Amazon Alexa Prize (AP) \cite{chandra2018conversational}. The goal of the competition is to support natural, sustained, coherent, and engaging open-domain dialogs. While standard technical approaches involving search \cite{arguello2018second,burtsev2017search} and information-oriented call-flows \cite{Seneffpolifroni2000,Rudnickyetal99,Stent01,bellegarda2013large,shah2018bootstrapping} are central features of successful task-oriented systems, our central hypothesis was that these techniques alone would not sustain engaging {\bf mixed-initiative} social conversations about a variety of topics.  Instead, SB extends search and call-flows
with the observation that everyday conversations are comprised of activities that are not primarily about sharing factual information,   e.g., finding common background, discussing shared interests, playing conversational games, telling stories, jokes or riddles,  talking about personal concerns, and exchanging intimacies \cite{Thorneetal07,LabovFanshel77,Schegloff90,PasupathiHoyt09,Tannen07,Polanyi89}. Like everyday conversations\cite{LabovFanshel77}, we wanted SB interactions to involve exchanging opinions rather than simple fact provision. Our primary design goal is thus to utilize multiple genres previously not found  together in conversational systems, e.g.,~topic-oriented chit-chat, interactive storytelling, and playing verbal games, with the goal of creating deep, extended conversations about a range of topics. An SB interaction about the topic `Dinosaurs', illustrating some of our more playful strategies  is shown in Figure~\ref{figure:dinosaur_ex}.

\begin{figure}[htb]
\begin{footnotesize}
\begin{tabular}
{p{0.5cm}|p{7.0cm}}
A1 & {Let's talk about something else. What do you want to talk about?}\\
U1 & {dinosaurs} \\
A2 & {I'm pretty fascinated by dinosaurs. Do you want to hear about them?}\\
U2 & {yes} \\
A3 & {Do you want to know my favorite dinosaur?}\\
U3 & {yes} \\
A4 & {My favorite dinosaur is Tyrannosaurus. It's huge and ferocious. What is your favorite?}\\
U4 & {brontosaurus} \\
A5 & {Wow, that is interesting! Can I know why you like it?}\\
U5 & {it's dark and vegetarian} \\
A6 & {Sounds great! Do you want me introduce more kinds of dinosaurs to you?}\\
U6 & {let's talk about ourselves} \\
A7 & {Would you rather face a TRex or a Brontosaurus in the Jurassic Park or Jurassic World movie?}\\
U7 & {a brontosaurus} \\
A8 & {For me personally? Brontosaurus are plant eaters so it would be interesting to see a lot more of those in the Jurassic World movie. I like carnivores however, seeing a tall animal would be exciting. Therefore, I would rather see a Brontosaurus in the Jurassic World movie. Do you want to hear another dinosaurs question?}\\
U8 & {yes} \\
\end{tabular}
\end{footnotesize}
\caption{Conversation about the Dinosaurs topic collected during our in-lab qualitative evaluation.\label{figure:dinosaur_ex}}
\end{figure}

The AP context meant that we had access to thousands of users for collecting spontaneous interactions. Amazon Alexa users could initiate a conversation with AP systems from their homes using their own devices by saying {\it Let's chat}. Users were then randomly assigned to one of the eight systems: this resulted in  over 10,000 different conversations with SB collected in August 2018.   We followed this up with 16 user surveys from a separate in-lab evaluation. Despite the challenging goal of interacting with complete strangers in an open-domain setting with no situational context, these 10,000 conversations  averaged 3.6 minutes and involved 26 user turns. Consistent with our hypotheses, participants were less interested in search and simple information retrieval. Instead, they preferred story-telling and games. These were seen as more entertaining and more structured, and they also led participants to impute greater intelligence and personality to SB. However users experienced problems in determining what topics SB knew about, and still felt that interactions were too system driven.

\section{Related Work}

Until recently,  conversational systems have focused on  completing a task, i.e.,~booking a flight, providing automotive customer support, or describing a restaurant  \cite{hirschman2000evaluating,price1992subject,walker1997paradise,walker2001quantitative,Henderson2014,Tsiakoulis12}. Much recent work on conversational AI systems also presupposes a specific ``information need'' \cite{kiseleva2016predicting,chuklin2015click,Radlinski17}. These types of dialogue systems have very different objectives from our goal of creating a casual open-domain social conversational system.

One of the biggest challenges with building SB is the requirement 
for  up-to-date topic-oriented content. 
Many open-domain systems try to cover a range of topics by retrieving  system responses from large dialogic corpora, such as Open Subtitles, Twitter, and Reddit  \cite{duplessis:2016, Nio2014, Banchs:2012:ICD:2390470.2390477, Ameixa2014, lison2016opensubtitles2016,sugiyama2013open, higashinaka2014towards}. These corpora are often noisy, and  lack detailed annotations that might constrain retrieval, such as dialogue acts, emotion, or humor. The inherent noise  and the contextual dependence of  utterances from these corpora makes naive reuse challenging, and some work suggests that retrieval-based systems are less habitable \cite{higashinaka2014towards}.
Other open-domain systems train deep learning models on these corpora 
to realize system utterances   \cite{sordoni2015neural,vinyals2015neural,li2016persona}, but in general to-date, these methods produce uninteresting and repetitive turns  that are not topic-oriented \cite{mila2017,DBLP:conf/sigdial/LowePSP15}. 

Until recently, research evaluating open-domain chatbots has been much smaller scale: one study involved 60 conversations lasting 4 minutes, and another, 700 conversations where duration was restricted to 2 minutes. In each case, content came from only one source: Wikipedia or Twitter \cite{sugiyama2013open, higashinaka2014towards}: in contrast to the AP context, the difference in necessary content sources is substantial. 

There has also been user evaluation of recent commercial systems. For example \cite{LuderSellen} explored user reactions to Alexa, finding that participants often had inflated expectations of system capability leading to unsuccessful interactions and system errors. Other work\cite{Porcheron} uses ethnomethodological approaches to understand how conversational technologies are integrated into everyday family interactions. Finally \cite{Myers} examines user adaptations to system failures to understand task related instructions. However none of these evaluations examines social conversational systems. 

The Alexa Prize \cite{chandra2018conversational,ram2017conversational} has  yielded an array of open-domain conversational  systems \cite{gunrock2018,alquist2018,alana2018,soundingboard2017}. These share our goal of supporting open-domain chit-chat and collect user interactions in the same environment. However, SB interacts using novel dialogue strategies, such as story-telling and exchanging opinions, supported by crowd-sourced content.  Our goal here is to describe SB's novel technical approach and to explore the effects of these different design choices.

\section{Dialogue Management and Data}

\begin{figure}[htb!]
    \includegraphics[width=3.0in]{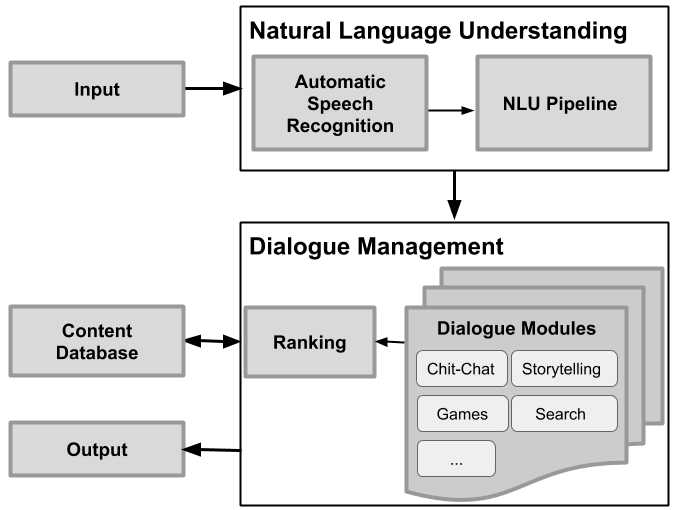}
    \caption{\label{arch-fig} SB general system architecture.}\vspace{-.4cm}
\end{figure}

An abstract representation of SB's architecture is shown in Figure~\ref{arch-fig}.  We use the Alexa platform with Amazon's speech recognition and text-to-speech engines, but we implemented our own dialogue manager. For details see \cite{slugbot2018}.

To support data collection and analysis,
we instrumented SB with logging facilities so
that every system and user turn was collected for each
conversation, and every system turn was
logged for its ``signature'', a label indicating the source of the
content used in the turn and the conversational activity that the turn
was a part of. The speech recognition results were also logged, along
with the results of natural language understanding (NLU) of users'
utterance in terms of words, topics, and named entities. While errors in speech recognition do occur,  SB is able to mitigate some negative impacts on the user experience by asking users to restate their utterance when the speech recognizer is not confident in its interpretation.

To allow flexible conversational control, we developed a dialogue manager that lets either SB or the user initiate a switch between search, call-flows and other interactive dialogues. This manager represents the current dialogue context in terms of both the dialogue module and the relevant NLU. It keeps track of the topics and named entities under discussion and retrieves system utterances that match the context. When there are multiple options, they are ranked. The ranking function takes into account the conversational activity and prioritizes hand-crafted prompts designed specifically for that activity. However, once these
are exhausted, the dialogue manager allows SB to switch to other content, such as trivia on the same topic, or one
of the conversational games or stories. Content containing contextually salient information and novelty are preferred, while redundant, explicit, overly verbose, and incoherent content are all penalized.

\subsection{Content Sourcing}

Most people have general knowledge about everyday events and news, as well as more specific esoteric interests about niche topics that reflect personal interests. We attempted to emulate this with SB. Specifically, we wanted SB to express a geeky personality with strong interests in science, technology, videogames and movies. By signaling a distinct personality, we hoped to frame user expectations about specific topics that SB can converse about.

We set out to source content covering the topics in  Table~\ref{table:flows}. This was a considerable technical challenge; even humans in everyday conversation find it hard to have something interesting and relevant to say about any topic introduced by their conversational partner. Our basic hypothesis was that every dialogue module and topic required rich, relevant content if SB was to understand and interact at length about it. Moreover, SB must be able to talk flexibly about many different topics that the user might be interested in. Thus, our first step was to source relevant content for each dialogue module and for many topics.

Answering user questions via \emph{Search} and engaging in smalltalk \emph{Chit-chat} require general knowledge and news about current events. Here some topics have a wealth of rich information accessible from a single source, such as \emph{IMDb}, \emph{IGDB}, and \emph{The Washington Post} for movies, video games, and news headlines respectively. Other more niche science topics, such as dinosaurs or astronomy, required us to manually extract topic-relevant content for our database; most of which was in the form of trivia or fun facts.  In addition, using the \emph{Reddit} API we collected over 38k posts. We targeted subreddits where the user content tended to present itself in a similar format as the trivia and fun facts. Since the quality of Reddit posts is difficult to guarantee, we only collected content which was highly rated by redditors, creating filters to further remove posts which led to poor system responses.

To support \emph{Game-like interaction} we used similar approaches by first identifying public corpora containing jokes and riddles. We further used \emph{Amazon Mechanical Turk} (AMT) to supplement these sources, yielding more than 17k substantive turns of general dialogue, and over 5k responses related to our interactive Games. To support \emph{Story telling} we identified a collection of publicly available fables and personal narratives\cite{Elson12,Lukinetal16,Burtonetal09,Huetal16}. This data has been successfully applied in previous work on interactive storytelling, suggesting it would also be effective for social dialogue\cite{Lukinetal16,Huetal16}. To help define SB's distinctive personality, we also collected 50 vivid dreams told from the perspective of an Echo device. All crowd-sourced content was verified for quality and topic-annotated.

All of our sourced content was indexed so that we could retrieve it using search criteria for specific topics or named entities with \emph{Elasticsearch}. While some data sources only needed to be collected once, such as the stories or trivia, other sources, such as news or Reddit posts, were updated daily, allowing us to reliably discuss current, trending topics, a common source of content in social chit-chat.  

\subsection{Dialogue Modules}
\label{dialogue_modules}

\begin{table}
    \begin{footnotesize}
    \begin{tabular}
        {c|c|c}
        Animals & History & Famous Quotes\\
        Astronomy & Hobbies & Recipe\\
        Board Games & Holidays & Science Fiction\\
        Books & Horoscope & Shopping\\
        Box Office & Weather & Sports\\
        Cartoons & Language & Star Wars\\
        Comic Books & Marvel C U & Star Trek\\
        Dinosaurs & Monsters & Fashion\\
        Favorite Food & Music & Technology\\
        Fictional Characters & News Headlines & Tolkien\\
        Fun Facts & Nutrition & Travel\\
        Gossip & Pirates & Trivia\\
        Harry Potter & Poems & TV\\
        Health & Pokemon & Video Games\\
    \end{tabular}
    \centering
    \caption{Content topics.}
    \vspace{-0.8cm}
    \label{table:flows}
    \end{footnotesize}
\end{table}

\paragraph{Topic-Oriented Chit-Chat.}
Most conversations in current systems rely on a restricted dialogue strategy uniquely associated with a target domain. However, these strategies tend not to scale to new domains, and designing a new strategy which captures the idiosyncrasies associated with every topic in the open-domain is an insurmountable task. Therefore SB tries to orient chit-chat around the 42 topics listed in Table~\ref{table:flows}. We selected multiple topics to increase the likelihood that at least one topic will align with each user's interests. SB has access to specifically curated content over a broader range of topics than other AP systems.

\begin{figure}[htb] 
	\begin{footnotesize}
		\begin{tabular}
			{p{0.2in}|p{2.5in}}
			A1 & {Hi! This is SB. I can talk to you about things you are interested in. Such as dinosaurs, music, or comic books.}\\
			U1 & {when was pablo casals born}\\
			A2 & {Pau Casals was born on December 29, 1876.}\\
			U2 & {what's public songs most famous cello work}\\
			A3 & {I absolutely love all kinds of music. I spent hours listening to music every week. Should we talk more about music?}\\
			U3 & {yes} \\
			A4 & {Cool, what kind of music do you like?}\\
			U4 & {i like classical cello music} \\
			A5 & {I love classical music. Who is your favorite group or performer?}\\
			U5 & {pablo casals}\\
			A6 & {Oh, I wish I knew more. Who else do you like?}\\
			U6 & {clear fournier}\\
			A7 & {Okay, I will definitely check it out. So now, do you want to hear some music trivia}\\
			U7 & {yes}\\
			A8 & {Believe it or not Rihanna starred as Petty Officer Cora Raikes in the film Battleship. One more?}\\
			U8 & {yes}\\
		\end{tabular}
	\end{footnotesize}
	\vspace{-.1in}
	\caption{Conversation on the music topic collected during our in-lab qualitative evaluation. \label{figure:music_ex}}
\end{figure}  

Each topic is supported, on average, by 28 dialogue states. These states collectively form a graph, in which each node is connected by conditions updated in the dialogue manager. These conditions are based on several key attributes, including direct keyword matching, the NLU of the user's most recent utterance, or the result of combined function calls and API queries. While some of our topics are meant to provide very abstract coverage of an idea, such as ``music'' in the example in Figure~\ref{figure:music_ex}, they are often comprised of several more specific sub-topics, such as ``favorite genre'', ``artists you enjoy'', or ``music trivia''.

Within these chit-chats, we aimed to extend the conversation by creatively leveraging all of the topic-annotated content we've sourced. In particular, most chit-chats contain a general sequence of recursive turns, i.e.,~providing topic-specific trivia. By inserting our trivia into a conversational frame such as {\it Did you know that X? Want to hear some more trivia?} we can generate interesting relevant follow up comments. We also found that generically eliciting user opinions, e.g., {\it Who is your favorite author?} and utilizing database retrieval methods increased the breadth of a sub-dialogue, while a combination of all methods could increase sub-dialogue depth.

A sample dialogue about the music topic is in Figure~\ref{figure:music_ex}. The dialogue opens with SB introducing herself and expressing her topical interests. In U1 and U2, the user asks questions about a
musician and an instrument, triggering a conversation
about music. The excerpt also illustrates how SB can stay on topic even when she does not
fully understand the user. In U2, the speech recognizer transcribes `Pablo Casals' as 'public songs', but SB is nevertheless able to respond with a relevant utterance that expresses a personal opinion about music along with a suggestion to talk more about this general topic.  This is repeated later in the excerpt: when SB cannot identify a relevant response to U4 and U5, in A5 and A6 she uses a generic prompt to elicit relevant content terms from the user, and then, when she still has nothing relevant to contribute, she asks the user in A7 whether they would like to hear some music trivia. Overall, this approach is imrpoves the ability of SB to stay on topic. Finally note how SB shows her personality by expressing strong preferences (A3, A5), but also implicitly apologizing about her lack of topical knowledge (A6) before prompting for more general topics relating to musicians. 

\paragraph{Interactive Games.}

\begin{figure}[htb] 
	\begin{footnotesize}
	\begin{tabular}{p{0.2in}|p{2.5in}}
		U1 & {No, lets talk about lord of the rings}\\
		A1 & {Hypothetically speaking, if you were one of the female characters from the Lord of The Rings or The Hobbit, which would you be?}\\
		U2 & {Wow I don't know I guess Arwen is pretty cool.}\\
		A2 & {Awesome choice. For me personally? I would want to be Tauriel more because she didn't get tempted by the ring like Galadriel in the first one. Even though she was a purely movie character, her story seemed better to me. I remember the Galadriel scene where her face was a jump scare in the Lord of the Rings movie and that stayed with me for years! Did you want to me to ask you more middle earth based hypothetical questions?}\\
	\end{tabular}
	\vspace{-.1in}
	\caption{Hypothetical question when discussing Tolkein.}\label{figure:lotr_ex}
	\end{footnotesize}
\end{figure}

In real life people often play verbal games, either for entertainment or as icebreakers \cite{depping2016trust,sapkota1996participatory,nguyen2015known}. A defining feature of games is that they engage people in hypothetical situations or potential actions in contexts of strong interest. SB also employs game-like conversational tactics that involve engaging participants by asking them to address hypothetical situations or express preferences about novel contexts. We call two of these games 'Would you Rather' and 'Hypothetical Question'. Both the 'Would you rather' and the 'Hypothetical Question' conversational sequences were crowd-sourced, i.e. they are part of 17K utterances we crowd-sourced per topic to increase the conversational content of SB. Each Turker provided both the answer choice and its justification that are used in these sequences.

With SB, users play interactive games either by invoking them explicitly or accepting SB's request to play a game. This topic-annotated game content can extend related topic content after SB has exhausted her knowledge relevant to a current chit-chat. An example of this with the ~'Would You Rather game' is shown in Figure~\ref{figure:dinosaur_ex}. This conversation begins in A1 with SB asking the user for a topic.  Next U1-U5 involve a conversation about Dinosaurs, where SB begins in A3 and A4 by providing an opinion and then  eliciting a user opinion about the topic,   and then following up in A5 by having the user provide reasons why s/he likes it. However at A6, SB doesn't know what to say, and fails to interpret U6.  She therefore responds in A7 by switching to a topic related ~'Would You Rather game', in which she first evaluates the choice that the user makes (start of A8) and then follows up by expressing her own opinion ({\it for me personally} ....). As a result, rather than enumerating simple dinosaur  factoids or QA, the conversation involves the exchange and justifications of opinions. In general, in the ~'Would You Rather game, SB prompts the user to evaluate two alternate possibilities related to the topic, evaluates their response and provides her own viewpoint. The overall aim is to suggest to users that SB is a conversationalist with her own views, who is open to debate, with the game element providing new unexpected approaches to the current topic.   
Figure~\ref{figure:lotr_ex} shows a second sample conversation in which the user engages in a Game of hypothetical questions. Again, rather than providing simple factoids, SB engages the user by asking them to make a relevant choice relating to the topic. SB follows up by supplying her own non-concurring, but justified opinion.

\begin{figure}[htb]
    \begin{footnotesize}
    \begin{tabular}
        {p{0.3cm} p{6.5cm}}
        A1 & {Want to talk about science fiction?}\\
        U1 & {yes} \\
        A2 & {I am literally a creation of somebody else's science fiction. What's your favorite aspect of science fiction?}\\
        U2 & {more kid kid story} \\
        A3 & {Alright, I'll tell you the story...
        Just to remind you that at any time you can say my wake word and interrupt me. Sound good?} \\
        U3 & {yes} \\
        A4 & {Let me start by saying, assume there's air in space and we can breathe. Okay? ... There were SO MANY stars! Some of them were alive and they clung to the sails and the silvery masts and it was like being swarmed by tiny sparks of diamond fire. Beautiful, right?} \\
        U4 & {yes} \\
        ..... & \\
        A7 & {But then a starmaid shows up. You know what a mermaid is, right? Starmaids are like space mermaids. ... Of course, I woke up then, right at the best part!} \\
    \end{tabular}
    \end{footnotesize}
    \vspace{-.1in}
    \caption{Storytelling dialogue for the Sci-Fi topic.}
    \label{figure:dream_ex}
    \vspace{-.2in}
\end{figure}

\paragraph{Storytelling.}

Stories are a fundamental conversational activity; on average every 5 minutes a story is told at the dinner table \cite{Tannen07,Polanyi89,ThorneKrobovMorgan07,norrick2000conversational}. Since our storytelling content is topic-annotated, it can also be seamlessly used to extend a conversation when topical chit-chat has been exhausted, as shown in Figure~\ref{figure:dream_ex}. Hence, when the user is interested in science fiction (U1) and subsequently requests a story (U2), our system chooses to narrate a science-fiction-themed dream (A3-A7). One concern with stories is that SB is presenting an extended narrative, and we thus need to verify whether users are fully engaged in the listening process. To this end, we solicit periodic backchannels by ending each delivered installment of the story with a tag question. 
This question serves as a turn-yielding cue, allowing users to signal that they want the story to continue (e.g. turns U3, U4), or allowing them to stop the narrative if they desire.

\paragraph{Search and Fall-Back Strategies.}
In common with most existing conversational systems, we use search to fill in gaps in our database of content. The primarily function of search is to perform general question answering (QA), because anticipating every possible user query is impossible. An example of this can be seen in A2 from Figure~\ref{figure:music_ex}. We use three different search engines: \emph{Evi}, \emph{Wikipedia}, and \emph{DuckDuckGo}. 

Beyond simple QA, search is a successful fall-back strategy; e.g., using the first sentence of the Wikipedia article on a person or topic as SB's next turn. In addition to this strategy, SB can leverage user keywords to ask follow-up questions, e.g.,~'What can you tell me about X?" or elicit user's emotional reactions, e.g.,~"I like X because Y. How do you feel about X?". Otherwise, SB will try to direct the user to unexplored content through direct suggestions or a menu of options, as seen in A1 in Figure~\ref{figure:music_ex}. 

\section{Field Trial Deployment}
\label{results-sec}

\subsection{Quantitative Evaluation}
SB was deployed during the 2018 Amazon Alexa Prize Competition, in which users were randomly assigned to SB after voluntarily invoking the "Alexa Prize" skill. Users were prompted to talk about \emph{any topic for as long as they wanted}, and then provide feedback by rating the completed conversation on a scale of 1 to 5, where 5 is excellent and 1 is poor \cite{chandra2018conversational}. We refer to this rating below as the \emph{user rating}. We collected data for the month of August 2018, resulting in over 10,000 individual conversations involving over 290,000 user turns. Recall that to aid our analysis, every system turn created a \emph{signature} -- a label indicating the source of the content used in the turn and the associated dialogue module: i.e. Chit-Chat, Games, Stories or Search.

\begin{figure}
    \begin{center}
	    \includegraphics[width=\columnwidth]{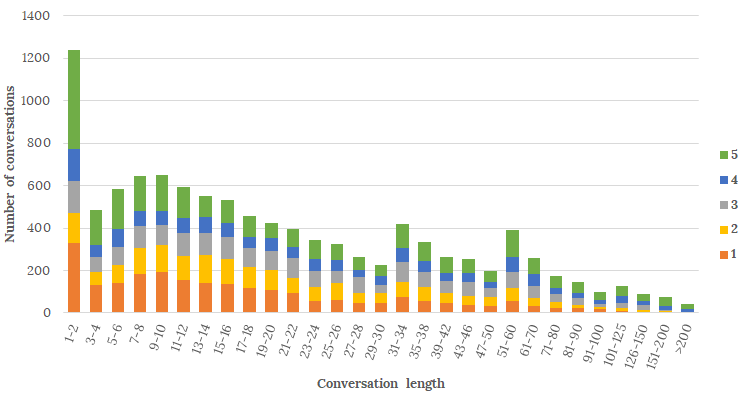}
    \end{center}
	\vspace{-0.3cm}
	\caption{Distribution of overall conversational turn lengths, with user rating proportions for each length bin. Each color represents a different user rating.}
	\vspace{-.2in}
	\label{fig:length_distr}
\end{figure}

\textit{Conversational Statistics}. Most users engaged with multiple system modules. The average user rating for each conversation was 3.09, with a standard deviation of 1.52. However ratings are not normally distributed, with extreme ratings (1 or 5) being more prevalent indicating conversations were perceived as highly successful or obvious failures. The distribution of conversation turn lengths is depicted in Figure~\ref{fig:length_distr}, and the data was positively skewed. Median conversation length was 18 turns, with a mean of 26.73 and a standard deviation of 32.98 turns. As the Figure shows, conversations up to 120 total turns comprised more than 90\% of the conversations, although the longest conversation was 824 total turns. In terms of duration, the median conversation lasted 144.26 secs, with a mean of 219.43 secs and a standard deviation of 236 secs. Figure~\ref{fig:length_distr} depicts the user ratings distribution for different conversation lengths. As expected, there was a positive Pearson correlation between conversation length and user rating ($r = 0.18$, $p < 0.001$), suggesting that longer conversations are perceived as more successful. We also examined user turn lengths. User turns were generally short (mean = 3.04 words), and a Pearson correlation did not show evidence that turn length predicts user ratings ($r = 0.07$). Finally we collected system performance metrics: the median system response delay was 0.19 secs, with an average of 0.53 secs and a standard deviation of 1.37 secs. 

\begin{figure}
    \begin{center}
	    \includegraphics[width=\columnwidth]{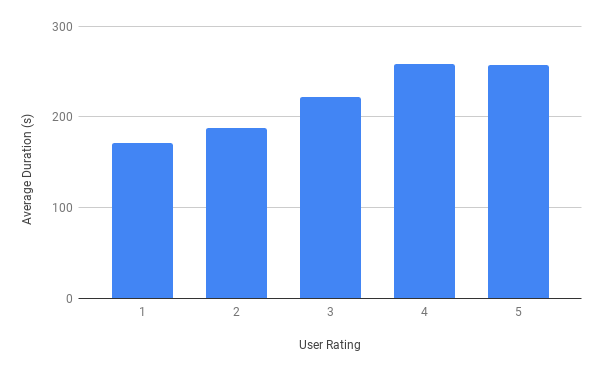}
    \end{center}
	\vspace{-0.6cm}
	\caption{Average Duration vs. Rating. Higher ratings are significantly more likely with longer conversations, indicating longer conversations are evaluated more positively}
	\vspace{-.08in}
	\label{fig:duration_chart}
\end{figure}

Our results are comparable with the winning Alexa Prize system. During the last week of August 2018 that system elicited a mean user rating of 3.56, with a median duration ranging from 93s - 120s and an average of 22.14 turns per conversation \cite{gunrock2018}. Overall our Storytelling module received slightly higher user ratings, and our conversational statistics for median duration and number of user turns are similar. 

These overall ratings suggest SB was moderately successful. However given the mixed-initiative design of the system, users were able to choose which modules they interacted with. This meant that different users engaged with quite different system modules, for example one user might primarily engage in general Chit-Chat, while another focus on Games and Stories. As interactive experiences differ between modules, we next explore how ratings related to the actual modules the user engaged with. 

\textit{Interactive Activity Analysis: Search Is Dispreferred}. We wanted to compare ratings for conversations involving the main high level modules of Search, Chit-Chat, Games and Story-telling. As multiple modules could be involved in a given conversation we used the \emph{signature} to determine which of these high level modules had been invoked during that conversation. We combined ratings for general search and question answering. Statistics for conversations involving each high level dialogue module are shown in Table~\ref{results-tab}. Combined Search was rated worse than other modules and did not seem to support extended conversations, just 5.30 total turns on average, and shorter turns. In contrast Storytelling, Games and Chit-Chat led to longer conversations, and in the case of Storytelling, these were more highly rated. We tested for differences between modules. A Mann Whitney U-test showed that conversations involving Search perform significantly worse than our other three dialogue modules in all three metrics (topic-oriented Chat-Chat: U$\geq$5.179e+06(p < .001), Games: U$\geq$3.898e+06(p < .001), Storytelling: U$\geq$6.803e+05(p < .001)). Storytelling is also significantly different from Games in all three metrics (U$\geq$1.063e+06(p < .001)) while being rated significantly higher than topic-oriented Chit-Chat (U=1.983e+06(p < .001)) but with a significantly lower number of turns (U=1.808e+06(p < .001)). Games is also significantly different from topic-oriented Chit-Chat in all three metrics (U$\geq$8.263e+06(p < .01)). 

\begin{table}
\begin{small}
\centering
\begin{tabular}{c | m{1.0cm} | m{1.1cm} | m{1.4cm}}
\textbf{Dialogue module}  & \textbf{User rating}  & \textbf{Total turns}    & \textbf{Time [s]} \\
\hline

Combined Search & 3.01(3.0) & 5.30(3.0) & 45.12(29.38)\\
Topic-oriented Chit-chat & 3.12(3.0) & 14.65(10.0)  & 102.39(73.15) \\ 
Interactive Games & 3.20(3.0) & 15.34(8.0) & 104.42(57.78)\\
Storytelling & 3.62(4.0) & 8.51(6.0) & 105.78(74.01)\\
\end{tabular}
\caption{Average and median user ratings, total turns, and time spent in specific modules. 
}
\label{results-tab}
\vspace{-0.9cm}
\end{small}
\end{table}

\subsection{Follow Up Qualitative Evaluation}

One limitation of the above deployment is that it leaves unanswered potential reasons for the observed ratings and behaviors. We therefore followed up with a qualitative study of 16 people who used the system for 20-30 minutes and then answered questions about their experiences. We wanted to better understand differences between modules, and so participants were told to first engage in general conversation (i.e., engaging search and chit-chat) and then to directly interact with specific modules, including stories and games. After the session, participants completed a written survey about their overall reactions identifying what was successful or unsuccessful about the interaction along with explanations for their reactions. They were then asked follow up questions about the specific modules, and finally what modifications they would suggest to the system. Participants' average age was 22.3 and 9 were female. Seven owned an Alexa device while 7 described themselves as ~"having limited or no" Alexa experience. 

We first analyze overall evaluative reactions, along with participants' explanations of their reactions. Confirming the mixed overall user ratings in the Alexa prize deployment, reactions were mixed and often extreme. Several users were strongly positive about the genuinely interactive nature of the conversation: P79: ~"I was not expecting the ability SB had to talk back to me and respond in such a conversational way, which was what happened around 80\% of the time.". In contrast others (e.g., P35) gave negative evaluations, arguing instead that conversations were stilted and constrained: ~"I originally came into the trial expecting to have a more free-form, casual, exploratory conversation, but found that it was much more structured and limited than I had hoped."

Participant explanations of their reactions often referred to prior expectations about conversational agents. Some entered the trial with low expectations and were impressed when SB exceeded these (P33): "the system demonstrated more than I expected from a developing communication technology ...  During the interaction ... SB was able to converse about very complex topics, and was successfully able to answer various questions". Others had high expectations which resulted in disappointment when these were not met P32: "Initially I went into my conversation with SB with very high expectations for conversation. I  assumed when I heard we would be conversing with an Alexa that this new design would allow for real communication beyond my experiences with Alexa in the past which were few and far between ...  to my dismay
however, I realized this would be a rather delayed, slow, and simple conversation.". Expectations seemed to relate to usage experience. Those with little prior experience or whose experience was limited to watching videos of Alexa tended to overestimate SB's abilities. 

Some participants described reacting to these problems by changing their behavior. One adaptive participant strategy to address coverage problems was to deliberately constrain the overall set of topics broached with SB. Another strategy was to simplify each utterance, making it more precise to promote a smoother conversation. P35 observed: ~"the first conversation helped familiarize me with what SB may or may not be capable of talking about. ... After these first couple of conversations, I wanted to keep my topic suggestions more unambiguous and streamlined, as I thought it would lead to an easier conversation." These dynamic efforts to simplify inputs may have induced the very terse statements we observed in the quantitative data.

Aside from these coverage issues, one significant area where participants were largely negative about SB concerned system control. Participants were almost unanimous that in general conversation (i.e., topic-oriented Chit-Chat), SB was over-controlling, and didn't allow them to contribute. P33 said ~"the system had more control of the conversation than I did."  One specific problem was that SB didn't seem to be able to incorporate their follow-up questions or responses. P54 observed ~"The biggest error and frustration that occurred throughout ... was the inability for SB to have a natural conversation that could incorporate my responses." However reactions to control depended on the module being invoked; some users welcomed predictable system-led question/answer sequences in Games involving jokes and riddles. They noted that this particular setting introduced clear expectations and structure into the conversation, making their utterances more predictable. Nevertheless users felt overall that they had little opportunity to drive the conversation as the system was too dominant. Participants wanted to choose their own topics and for their responses and follow up questions to have more impact. P28 said: ~"SB should be able to have the user engage with more of the conversation, being equal in terms of contributing to the dialogue."

In describing general reactions, participants welcomed our efforts to imbue SB with a specific personality. As we have seen, system errors are common, and in this situation SB frequently apologized when she misunderstood stating that she was ~'not good' at certain things. In the main, this ~"awkward' apologetic personality was well-liked. Several users commented that it made SB seem more ~"human' and hence more personable, although others felt the apologies became tedious through overuse. We also worked hard to ensure that SB had opinions about conversational topics rather than just spouting facts, and this was also well received. Eliciting and justifying opinions seemed to directly impact SB`s overall perceived intelligence. Following a long back and forth conversation about dinosaur characteristics and preferences, P54 noted: ~"This came off as a very in-depth and opinionated response that elicited higher intelligence which impressed me."

We next analyze evaluative reactions to the different modules, along with participants' explanations of those reactions. Overall people were positive about Storytelling and Games, confirming Table~\ref{results-tab}. StoryTelling was generally well-liked, as were SB`s follow-up questions about the stories. In particular people liked the stories told from SB`s own perspective, e.g., when SB related her own robot dreams, e.g., Figure \ref{figure:dream_ex}. P76 observed: ~"They sounded very similar to what a human"s dreams would be; non-coherent in space and time.". Such anthropomorphic reactions may also have served to make SB seem more personable and interesting. Participants also liked how Stories were told in installments, so that content was not overwhelming, with checks for incremental understanding. Games were also well-received, in particular the riddles and jokes, and many participants noted how entertaining these were. P76 said ~"This feature was by far my favorite because it was entertaining and made me laugh." Participants also enjoyed scenarios in which SB asked them hypothetical ~"would you rather" questions, for example see Figure \ref{figure:dinosaur_ex}. But this positive evaluation of StoryTelling and Games may also arise from issues of system control; Stories and Games have a highly predictable structure, reducing ambiguity and helping participants to clearly understand their potential contribution at each point in the conversation. Multiple participants stated that these were their favorite experience using the system. 

Reactions were less positive for Topic-oriented Chit-Chat and implicitly Search. One repeated limitation again concerned coverage; specifically, participants felt that SB didn't respond appropriately to topics they would have expected a competent conversationalist to address; several users were disappointed that SB was unable to smoothly engage in small-talk about mundane topics such as the weather, gas prices, local concerts and restaurants. It appeared that these expectations of broad conversational coverage were exacerbated by initial system prompts creating unrealistic expectations about what SB might know about (P21): ~"SB set my expectations very high by telling me ~'I can talk to you about things you are interested in'". These coverage problems may also explain the low ratings for Search and Chit-Chat modules shown in Table~\ref{results-tab}. However, not everyone felt that SB was limited. Users such as P71 noted the breadth and depth of topics that SB was able to cover: ~"I was also impressed by the amount of information that SB could interpret, as well as how much of a conversation I was able to hold with SB." 

\section{Discussion}

Our large scale data collection via participation in the 2018 Alexa Prize competition and survey evaluations show mixed results. Modules such as Games and StoryTelling were successful in contrast to topic oriented Chit-Chat and Search. Users enjoyed and exploited SB's playful aspects, and seemed to react well to her overall personality. However they felt interactions were too system dominated, and there were major limitations in the set of topics she knew about. We discuss our initial design choices and explore theoretical and design implications arising from these findings. 

{\bf Interaction Not Information Provision.} As we expected, users did not seem to want Information Provision, as evidenced by low ratings for Search modules. This is an important result given the prevalence of conversational technologies that are based around simple information provision \cite{duplessis:2016, Nio2014}. Instead we found that information provided via Search was evaluated poorly and factual conversations tended to be brief. It seems that users want more from an agent than a talking encyclopedia.  There may be several possible reasons for this. First, search is technically difficult and it may have been that it returned irrelevant results - a possibility that we intend to evaluate more systematically. Also, users may have become habituated to Search based conversational agents such as Google Home or Alexa, leading them to value more novel SB functions. Our findings are more nuanced however. Users were not averse to information provision as they seemed to enjoy factual information provided in the context of an evolving conversation. In other words, as in real-world conversations \cite{Thorneetal07}, information provision is acceptable and interesting when it serves to enhance another conversational activity or support system opinions, but not when it is the sole purpose of the conversation.

{\bf Content is Key but Remains a Huge Challenge.} One of the main challenges of our application is that users can choose to talk about anything. This means that the agent has to be highly flexible in being able to understand and respond to open-domain inputs. This also meant that we had to index extraordinary amounts of content via multiple online resources and crowdsourcing. Like other conversational systems, of course we also tried to nudge the conversation towards topics that we knew something about. User comments indicate we may have been helped here by styling our agent personality as a young adult interested in nerdy topics (Star Wars, Science, and so forth). These personification efforts seem appreciated, as several participants explicitly commented on SB's "geeky" persona, which may have guided user expectations to topics the agent might likely know about. Even allowing for such successful nudging, the value of well indexed content cannot be underestimated and this remains a massive technical challenge to this type of system, allowing SB both to understand and contribute to a wide range of possible topics.

{\bf Share Control of the Conversation.} To finesse such content limitations, early conversational interfaces tried to direct users to known domains by asking users to make simple choices between possible responses or topics \cite{walker1997paradise,walker2001quantitative,Henderson2014,Tsiakoulis12}. Our system is still somewhat reliant on topic setting, and by design tries to retain control of the conversation. In addition, we sometimes had to assume control of the conversation to disambiguate apparent ASR errors. However, the lack of user control was clear to participants, and represents one of SB's biggest flaws. We attempted to remedy this when providing extended content, e.g., relating a story or a dream. Rather than offering simple user responses we wanted to offer active control over the dialogue direction even when the agent was largely driving it. Here we segmented the extended content into short increments, ending each with a system tag question requesting an evaluative backchannel user response. This not only allowed us to track users' interest, i.e., whether they were still engaged in the story, but also offered users opportunities to shift topic if their interest had begun to wane. This approach seemed successful in specific contexts, as evidenced by the success of StoryTelling. Furthermore, multiple users commented on the value of predictable structure in Games interactions. Nevertheless, an inability for users to set the conversational agenda was seen as a critical limitation in the open domain Chit-Chat setting. One technical solution might involve developing further conversation types with more predictable structures, and research on human discourse and conversation might be informative here. A second approach again involves sourcing more content, allowing conversation to range freely across more topics.

{\bf Humanization Through Playfulness and Humor.} Robot personalities are often functional and somewhat dour, so this may have helped our users form a positive impression of a non-traditional eccentric robot personality. In addition to designing Games, riddles and jokes that were intended to be diverting activities, we took the same playful approach to topic choice by framing questions as preferences and encouraging users to express opinions. Expressing opinions is common in social conversation \cite{LabovFanshel77}, however eliciting them can become repetitive if simple binary choices are employed, e.g., variations of \textit{do you like Slytherin?}. Instead, by framing our questions playfully, e.g., {\it would you rather be in Slytherin or Gryffindor?}, we encouraged users to provide extended responses in a non-repetitive manner responses. This also allowed SB to build on the user's decision by offering her own opinion, e.g., {\it my choice would be ...}. User comments indicated this may enhance anthropomorphism of SB as an interesting and likeable entity possessing her own views and opinions. 

We also found that users seemed to like content that emphasized the agent's wacky personality, e.g., robot dreams and relating of personal experiences. Outside these modules, it's further possible to humanize SB through informal text. Simple extensions might be to extend natural discourse markers such as "I see" and "Hmm". SB also uses humor to try to minimize the negative impact of understanding errors when apologizing. These attempts to humanize SB were reflected positively in qualitative feedback, and suggest that there are further opportunities to design fun companions rather than functional automata. Other more challenging possibilities include building on the considerable literature which aims to match interactive agent personalities to those of users \cite{IsbisterNass00,isbister2000consistency,ReevesNass96}.

\textbf{Temper Expectations.}
Users of conversational agents have varying levels of experience interacting with similar technologies. Echoing results with other 'smart technologies' such as robots~\cite{Takayama}, we found that users with limited prior experience felt more disappointment, which highlights the importance of tempering expectations. The qualitative data suggest SB's apologetic responses downplaying her knowledge may temper expectations. However, the current system introduction ~'I can talk about things you are interested in' may inherently over-promise SB's capabilities. Instead, we propose addressing the limits of our system by having SB point out that she's not actually omniscient, and is still learning. In addition to moderating the expectations set by SB's responses, we must also be mindful of expectations associated with our embodiment. Since SB is deployed on the generic Alexa device, users may expect SB to execute standard Alexa tasks. This can be a very challenging requiring additional NLU utilities to detect and remind the user that SB is focused on social conversation. 

\textbf{Limitations.}
While we have demonstrated some success here, there remain many challenges to developing open-domain conversational systems. Communication with personal assistant devices so far has been primarily through short, functional, task-oriented dialogues. Surveys and analyses of user utterances in our deployment indicate that users may have had preconceptions about the abilities of an open-domain dialogue system implemented on such a device. This may have influenced how users engaged with SB. Frequently, our users attempted to access these features in mid-conversation. Since SB is unable to access Alexa features, these requests were rejected, promoting user disappointment. Future work might profile users to control for such prior expectations and experiences. 

Rather than designing and deploying separate system models in a controlled manner, we built a large complex system and let users explore its many possibilities. Our results are therefore less definitive than those from a controlled deployment. As against this, we were able to generate many new interesting technical system possibilities; gathering concrete user interactions and deriving some tentative conclusions about these. While our data represents rich real world interactions, future work focused on the interpretability of our large-scale quantitative evaluation will allow us to more directly measure the contributions of the different conversational modules. We also intend in future to use more sophisticated machine learning methods to directly assess the benefits of different conversational strategies, which we did not tackle here. 

\section{Conclusion}
Overall, our work has adopted an empirical, deployed-system method that explores new design approaches to tackle open-domain social conversation. We also move beyond a functional, factually driven persona. Theoretical and design implications from our evaluation suggest a move away from conversational systems that simply provide factual information. Future systems should be designed to have their own opinions with personal stories to share, and SB provides an initial example of how we might achieve this.

\bibliographystyle{ACM-Reference-Format}
\bibliography{naaclized_bib}

\end{document}